\documentclass{article}
\usepackage{spconf}
\usepackage[bookmarks=false,hyperfigures=false,hidelinks=true]{hyperref}
\usepackage{amsmath}
\usepackage{amssymb, subfigure, url, doi, xspace}
\usepackage{mathtools}
\ninept

\usepackage{ifpdf}
\ifpdf
  \usepackage[final]{graphicx}
  \DeclareGraphicsExtensions{.pdf,.png}
\else
  \usepackage[final]{graphicx}
  \DeclareGraphicsExtensions{.pstex,.ps,.eps,.png}
\fi
\usepackage{color}

\newcommand{\ie}{i.e.\xspace}

\newcommand{\eq}[1]{(\ref{eq:#1})}
\newcommand{\fig}[1]{Fig.~\ref{fig:#1}}

\newcommand{\tbl}[1]{Table~\ref{tbl:#1}}
\newcommand{\sctn}[1]{Sec.~\ref{sec:#1}}

\newcommand{\smallmath}[1]{%
  {\small
    \setlength{\abovedisplayskip}{6pt}
    \setlength{\belowdisplayskip}{\abovedisplayskip}
    \setlength{\abovedisplayshortskip}{0pt}
    \setlength{\belowdisplayshortskip}{3pt}
    #1
  }%
}%

\newcommand{\resizemath}[2]{%
  \resizebox{#1}{!}{%
    $ \displaystyle #2 $%
  }%
}%

\makeatletter%
\if@twocolumn%
\else%
\renewcommand{\resizemath}[2]{%
  #2
}%
\fi
\makeatother

\allowdisplaybreaks[2]

\newcommand{\mb}[1]{\mathbf{#1}}
\newcommand{\mbs}[1]{\boldsymbol{#1}}

\newcommand{\sign}{\mathop{\mathrm{sign}}}

\newcommand{\norm}[1]{\left\| #1 \right\|}
\DeclarePairedDelimiterX{\normsz}[1]{\lVert}{\rVert}{#1}
\newcommand{\abs}[1]{\left| #1 \right|}

\DeclareMathOperator*{\argmin}{arg\,min}

\date{\today}

\begin{document}

\title{Convolutional Sparse Representations with Gradient Penalties}

\name{Brendt Wohlberg\thanks{This research was supported by the
    U.S. Department of Energy via the LANL/LDRD Program.}}
\address{Theoretical Division\\
  Los Alamos National Laboratory\\
  Los Alamos, NM 87545, USA}

\maketitle

\begin{abstract}
While convolutional sparse representations enjoy a number of useful properties, they have received limited attention for image reconstruction problems. The present paper compares the performance of block-based and convolutional sparse representations in the removal of Gaussian white noise. The usual formulation of the convolutional sparse coding problem is slightly inferior to the block-based representations in this problem, but the performance of the convolutional form can be boosted beyond that of the block-based form by the inclusion of suitable penalties on the gradients of the coefficient maps.
\end{abstract}


\section{Introduction}

\emph{Sparse representations} are well-established as a tool for inverse problems in a wide variety of areas, including signal and image processing, computer vision, and machine learning~\cite{mairal-2014-sparse}. The standard form is a linear representation $D \mb{x} \approx \mb{s}$, where $D$ is the \emph{dictionary}, $\mb{x}$ is the representation, and $\mb{s}$ is the signal to be represented. When $D$ is a linear transform with a fast transform operator, such as the Discrete Wavelet Transform, these representations can be computed for large images, but when $D$ is learned from training data and represented as an explicit matrix, this is not feasible, the standard approach being to independently compute the representations over a set of overlapping image patches. \emph{Convolutional sparse representations} are a recent\footnote{More accurately, the label \emph{convolutional} is recent, but the equivalent \emph{translation invariant sparse representations} are much older~\cite[Sec. II]{wohlberg-2016-efficient}.} alternative that replace the general linear representation with a sum of convolutions\footnote{Typically circular convolutions~\cite{wohlberg-2016-boundary}.} $\sum_m \mb{d}_m \ast \mb{x}_m \approx \mb{s}$, where the elements of the dictionary $\mb{d}_m$ are linear filters, and the representation consists of the set of coefficient maps $\mb{x}_m$.

There is growing interest in imaging and image processing applications of the convolutional form~\cite{gu-2015-convolutional, liu-2016-image, wohlberg-2016-convolutional2, zhang-2016-convolutional, quan-2016-compressed, zhang-2017-convolutional}.  Surprisingly, denoising of Gaussian white noise, arguably the simplest of all imaging inverse problems, has received almost no attention beyond a very brief example providing insufficient detail for reproducibilty~\cite[Sec. 4.4]{zeiler-2010-deconvolutional}. The present paper argues that, despite its numerous advantages in many contexts, the convolutional form is not competitive for the Gaussian white noise denoising problem, but that these deficiencies can be mitigated by moving beyond simple $\ell_1$ regularization, the specific form being investigated here consisting of additional penalties on the gradients of the coefficient map\footnote{A weighting strategy applied to the $\ell_1$ penalty has also been found to improve the denoising performance of convolutional sparse representations~\cite[Sec. 8]{wohlberg-2017-convolutional3}, but that approach is not considered here due to space constraints.}.

It is emphasised that these extensions have relevance beyond the specific denoising test problem considered here, in that the improved performance reported on this problem can also be expected to have an impact on more general image reconstruction problems, e.g. when convolutional sparse coding is employed as the prior within the plug-and-play priors framework~\cite{venkatakrishnan-2013-plugandplay2, sreehari-2016-plug}. There is also evidence that the inclusion of such gradient penalties enhances the performance of convolutional sparse representations in certain image decomposition/restoration problems~\cite{zhang-2016-convolutional, zhang-2017-convolutional}.

\section{Convolutional Sparse Coding}
\label{sec:cbpdn}

The most widely used form of convolutional sparse coding is Convolutional Basis Pursuit DeNoising (CBPDN), defined as
\vspace{-1.2mm} {\small
\begin{equation}
\argmin_{\{\mb{x}_m\}} \frac{1}{2} \normsz[\Big]{\sum_m \mb{d}_m \ast \mb{x}_m
- \mb{s}}_2^2 + \lambda \sum_m \alpha_m \norm{\mb{x}_m}_1 \; ,
\label{eq:convbpdn}
\vspace{-1.2mm}
\end{equation}
} where the $\alpha_m$ allow distinct weighting of the $\ell_1$ term for each filter $\mb{d}_m$.  At present, the most efficient approach to solving this problem~\cite{wohlberg-2016-efficient} is via the Alternating Direction Method of Multipliers (ADMM)~\cite{boyd-2010-distributed} framework. An outline of this method is presented here as a basis for extensions proposed in following sections.

Problem ~\eq{convbpdn} can be written  as
\vspace{-1.2mm}
{\small
\begin{equation}
\argmin_{\mb{x}} \; (1/2) \normsz[\big]{D \mb{x}
- \mb{s}}_2^2 + \lambda \norm{\mbs{\alpha} \odot \mb{x}}_1 \; ,
\label{eq:convbpdnblk}
\vspace{-1.2mm}
\end{equation}
} where $\odot$ is the Hadamard product, $D_m$ is a linear operator such that $D_m \mb{x}_m = \mb{d}_m \!\ast\! \mb{x}_m$, and $D$, $\mbs{\alpha}$, and $\mb{x}$ are the block matrices/vectors \vspace{-1mm}
\begin{equation}
\resizemath{.9\hsize}{
D = \left( \begin{array}{ccc} D_0 & D_1 &
    \ldots \end{array} \right)
\;\;\;
\mbs{\alpha} = \left( \begin{array}{c}  \alpha_0 \mb{1}\\ \alpha_1 \mb{1}\\
    \vdots  \end{array} \right)
\;\;\;
\mb{1} = \left( \begin{array}{c}  1\\ 1\\
    \vdots  \end{array} \right)
\;\;\;
\mb{x} = \left( \begin{array}{c}  \mb{x}_0\\ \mb{x}_1\\
    \vdots  \end{array} \right)
\;.
}
\vspace{-1mm}
\label{eq:dxcbpdn}
\end{equation}
This problem can be expressed in ADMM standard form as
\vspace{-1mm}
\begin{equation}
\argmin_{\mb{x},\mb{y}} \; (1/2) \normsz[\big]{
    D \mb{x} - \mb{s}}_2^2 \!+\! \lambda
  \norm{\mbs{\alpha} \odot \mb{y}}_1 \text{ s.t. } \mb{x} \! - \!
  \mb{y} \!=\! 0 \; ,
\label{eq:cbpdnsplit}
\vspace{-1mm}
\end{equation}
which can be solved via the ADMM iterations
{\small
  \setlength{\abovedisplayskip}{6pt}
  \setlength{\belowdisplayskip}{\abovedisplayskip}
  \setlength{\abovedisplayshortskip}{0pt}
  \setlength{\belowdisplayshortskip}{3pt}
\begin{align}
  \mb{x}^{(j+1)} &= \argmin_{\mb{x}} \frac{1}{2}
  \normsz[\big]{D \mb{x} - \mb{s}}_2^2 +
  \frac{\rho}{2}  \norm{
    \mb{x} - \mb{y}^{(j)} + \mb{u}^{(j)}}_2^2 \label{eq:bpdnxprob} \\[-1pt]
  \mb{y}^{(j+1)} &= \argmin_{\mb{y}} \lambda
  \norm{\mbs{\alpha} \odot \mb{y}}_1 +
   \frac{\rho}{2}
 \norm{ \mb{x}^{(j+1)} - \mb{y} +
    \mb{u}^{(j)}}_2^2 \label{eq:bpdnyprob}  \\[-1pt]
  \mb{u}^{(j+1)} &= \mb{u}^{(j)} + \mb{x}^{(j+1)} -
  \mb{y}^{(j+1)} \; . \label{eq:bpdnuprob}
\end{align}
}%

The solution to~\eq{bpdnyprob} is given by the soft thresholding operation~\cite[Sec. 6.5.2]{parikh-2014-proximal}
$\mb{y} = \sign(\mb{z}) \odot \max(0, \abs{\mb{z}}  - \lambda \mbs{\alpha} / \rho)$
where $\mb{z} = \mb{x} + \mb{u}$.
The only computationally expensive step is~\eq{bpdnxprob}, which can be solved via the equivalent DFT domain problem
\vspace{-1.2mm} \smallmath{
\begin{equation}
\argmin_{\hat{\mb{x}}} \; (1/2) \normsz[\big]{
 \hat{D} \hat{\mb{x}} - \hat{\mb{s}} }_2^2 + (\rho/2) \norm{
 \hat{\mb{x}} - \left( \hat{\mb{y}} - \hat{\mb{u}} \right) }_2^2 \;,
\label{eq:linprobdft}
\vspace{-0.8mm}
\end{equation}
}%
where $\hat{\mb{z}}$ denotes the DFT of variable $\mb{z}$.
The solution for~\eq{linprobdft} is given by the $MN \times MN$ linear system (for $M$ filters and an image $\mb{s}$ with $N$ pixels)
\vspace{-1.2mm}
\begin{equation}
  (\hat{D}^H \hat{D} + \rho I) \hat{\mb{x}} = \hat{D}^H \hat{\mb{s}} +
  \rho \left(\hat{\mb{y}} - \hat{\mb{u}} \right)
  \; .
\label{eq:dhdrhoi}
\vspace{-0.8mm}
\end{equation}
The key to solving this very large linear system is the observation that it can be decomposed into $N$ independent $M \times M$ linear systems~\cite{bristow-2013-fast}, each of which has a system matrix consisting of the sum of rank-one and diagonal terms so they they can be solved very efficiently by exploiting the Sherman-Morrison formula~\cite{wohlberg-2014-efficient}.

\section{Gradient Regularization}
\label{sec:tvreg}

An extension of~\eq{convbpdn} to include an $\ell_2$ penalty on the gradients of the coefficient maps was proposed in~\cite{wohlberg-2016-convolutional2}. The primary purpose of this extension was as a regularization for an impulse filter intended to represent the low-frequency components of the image, but a small non-zero regularization on the other dictionary filters was found to provide a small improvement to the impulse noise denoising performance~\cite{wohlberg-2016-convolutional2}. Considering the edge-smoothing effect of $\ell_2$ gradient regularization, a reasonable alternative to consider is Total Variation (TV) regularization.  We consider three different variants:
\begin{enumerate}
\setlength{\itemsep}{0.7pt}
\item scalar TV~\cite{rudin-1992-nonlinear} applied
  independently to each coefficient map,
\item vector TV~\cite{blomgren-1998-color} applied jointly to the set of
  coefficient maps,
\item scalar TV~\cite{rudin-1992-nonlinear} applied to the
  reconstructed image components $D_m \mb{x}_m$ rather than to the coefficient maps $\mb{x}_m$.
\end{enumerate}

\subsection{Scalar TV on Coefficient Map}
\label{sec:sclrtv}

The CBPDN problem extended by adding a scalar TV term on each
coefficient map can be written as
\vspace{-1.5mm}
\begin{align}
   \argmin_{\{\mb{x}_m\}} \;& \frac{1}{2} \normsz[\Big]{\sum_m \mb{d}_m \ast
     \mb{x}_m - \mb{s}}_2^2 + \lambda \sum_m \alpha_m \norm{\mb{x}_m}_1 +
   \nonumber \\& \mu \sum_m \beta_m \norm{
 \sqrt{(\mb{g}_0 \ast \mb{x}_m)^2 + (\mb{g}_1  \ast \mb{x}_m)^2}}_1
\; ,
\label{eq:convl1grd}
 \vspace{-1mm}
\end{align}
where $\mb{g}_0$ and $\mb{g}_1$ are filters that compute the gradients along image rows and columns respectively.  The TV term can be written as $\mu \sum_m \beta_m \norm{\sqrt{(G_0 \mb{x}_m)^2 + (G_1 \mb{x}_m)^2}}_1$ where linear operators $G_0$ and $G_1$ are defined such that $G_l \mb{x}_m = \mb{g}_l \ast \mb{x}_m$, and defining\footnote{Note that the $\Gamma_l$ notation is overloaded, taking on a different definition in each section.}  \vspace{-1mm}
\begin{equation}
\resizemath{.45\hsize}{
\Gamma_l = \left( \begin{array}{ccc}
    \beta_0 G_l & 0 & \ldots \\
    0 & \beta_1 G_l  & \ldots \\
    \vdots & \vdots & \ddots
  \end{array} \right)
}
\vspace{-1mm}
\end{equation}
allows further reduction to $\mu \norm{\sqrt{(\Gamma_0 \mb{x})^2 + (\Gamma_1 \mb{x})^2}}_1$.

Problem~\eq{convl1grd} can be written in standard ADMM form as
\begin{align}
\hspace{-2mm}
\argmin_{\mb{x},\mb{y}_0,\mb{y}_1,\mb{y}_2} \frac{1}{2}
\normsz[\big]{D\mb{x} - \mb{s}}_2^2 &+ \lambda
\norm{\mbs{\alpha} \odot \mb{y}_2}_1 +
\mu \normsz[\Big]{\sqrt{\mb{y}_0^2 + \mb{y}_1^2}}_1 \nonumber \\
& \mkern-36mu  \text{ s.t. }
  \left( \begin{array}{c} \Gamma_0 \mb{x} \\ \Gamma_1 \mb{x} \\
     \mb{x} \end{array} \right)
  -
 \left( \begin{array}{c} \mb{y}_0 \\ \mb{y}_1 \\ \mb{y}_2 \end{array} \right)
 = 0
  \; .
\label{eq:cbpdnsplitmd}
\end{align}
The resulting $\mb{x}$ subproblem has the form \vspace{-1mm}
\begin{align}
   \argmin_{\mb{x}} \;& \frac{1}{2} \norm{D
     \mb{x} - \mb{s}}_2^2 +
   \frac{\rho}{2} \norm{\Gamma_0 \mb{x} - \mb{y}_0 + \mb{u}_0}_2^2  +
   \nonumber \\[-2pt] &
   \frac{\rho}{2} \norm{\Gamma_1 \mb{x} - \mb{y}_1 + \mb{u}_1}_2^2 +
   \frac{\rho}{2} \norm{\mb{x} - \mb{y}_2 + \mb{u}_2}_2^2 \; ,
\label{eq:cbpdngrdl1prob}
 \vspace{-2mm}
\end{align}
and the solution of the equivalent DFT domain problem is given by
\vspace{-1.5mm}
\begin{align}
  (\hat{D}^H \hat{D} + &\rho I + \rho \hat{\Gamma}_0^H \hat{\Gamma}_0
  + \rho \hat{\Gamma}_1^H \hat{\Gamma}_1) \hat{\mb{x}} =
  \hat{D}^H \hat{\mb{s}} + \rho \left(\hat{\mb{y}}_2 - \hat{\mb{u}}_2
    + \vphantom{\hat{\Gamma}_0^H}  \right. \nonumber \\[-2pt]
 & \left.  \hat{\Gamma}_0^H (\hat{\mb{y}}_0 - \hat{\mb{u}}_0)   +
      \hat{\Gamma}_1^H (\hat{\mb{y}}_1 - \hat{\mb{u}}_1)
\right)   \; .
\label{eq:tvxprob}
\vspace{-2mm}
\end{align}
Since $\hat{\Gamma}_0^H \hat{\Gamma}_0$ and $\hat{\Gamma}_1^H \hat{\Gamma}_1$ are diagonal (the $\hat{G}_l$ are diagonal, and therefore so are $\hat{\Gamma}_l$), they can be grouped together with the $\rho I$ term; the independent linear systems described in~\sctn{cbpdn} are again composed from rank-one and diagonal terms and the Sherman-Morrison solution~\cite{wohlberg-2014-efficient} can be directly applied without any substantial increase in computational cost.

The $\mb{y}$ subproblem for~\eq{cbpdnsplitmd} can be decomposed into the independent problems
\vspace{-1.5mm}
\begin{align}
\argmin_{\mb{y}_2} \;& \lambda \norm{\mbs{\alpha} \odot \mb{y}_2}_1 + (\rho / 2) \norm{\mb{x} - \mb{y}_2 + \mb{u}_2}_2^2 \label{eq:tvy2prob} \\[-4pt]
\argmin_{\mb{y}_0, \mb{y}_1} \;& \mu \normsz[\Big]{\sqrt{ \mb{y}_0^2 + \mb{y}_1^2}}_1 + (\rho / 2) \norm{\Gamma_0 \mb{x} - \mb{y}_0 + \mb{u}_0}_2^2 \nonumber \\[-4pt]  & \hspace{15ex} + (\rho / 2) \norm{\Gamma_1 \mb{x} - \mb{y}_1 + \mb{u}_1}_2^2 \;.
\label{eq:tvy01prob}
\vspace{-2mm}
\end{align}
The solution for~\eq{tvy2prob} is the same as that for~\eq{bpdnyprob}, and~\eq{tvy01prob} can be solved by use of the block soft thresholding operation~\cite[Sec. 6.5.1]{parikh-2014-proximal} applied in the same way as in the ADMM algorithm for the standard isotropic TV denoising problem~\cite{wang-2008-new, yang-2009-fast},~\cite[Sec. 4.1]{goldstein-2009-split}, i.e.  \vspace{-3mm}
\begin{align}
\mb{y}_l = \frac{\mb{z}_l}{\sqrt{\mb{z}_0^2 + \mb{z}_1^2}} \max\Big(0, \sqrt{\mb{z}_0^2 + \mb{z}_1^2} - \frac{\mu}{\rho}\Big) \quad l \in \{0,1\}
\label{eq:l2shrnksclr}
\vspace{-1mm}
\end{align}
where $\mb{z}_l = \Gamma_l \mb{x} + \mb{u}_l$ for $l \in \{0,1\}$.

\subsection{Vector TV on Coefficient Maps}
\label{sec:vectv}

Instead of independently applying scalar TV to each coefficient map, one can treat the set of coefficient maps as a multi-channel image and apply Vector TV~\cite{blomgren-1998-color}, originally designed for restoration of colour images.  The corresponding extension of the CBPDN problem can be written as \vspace{-1mm}
{\small
\begin{align}
   \argmin_{\{\mb{x}_m\}} \;& \frac{1}{2} \normsz[\Big]{\sum_m \mb{d}_m \ast
     \mb{x}_m - \mb{s}}_2^2 + \lambda \sum_m \alpha_m \norm{\mb{x}_m}_1 +
   \nonumber \\& \mu   \normsz[\Big]{ \sqrt{\vphantom{\sum}  \smash[b]{\sum_m \beta_m \left[
    (\mb{g}_0 \ast \mb{x}_m)^2 + (\mb{g}_1  \ast  \mb{x}_m)^2 \right] }}}_1
\; .
\label{eq:convl1grdv}
 \vspace{-1mm}
\end{align}
}
Using the $G_l$ as defined in~\sctn{sclrtv}, the TV term can be written as
\[
\resizemath{.55\hsize}{
\mu \normsz[\Big]{\sqrt{\vphantom{\sum} \smash[b]{\sum_m \beta_m \left[
(G_0 \mb{x}_m)^2 + (G_1 \mb{x}_m)^2 \right] }}}_1
} \;.
\]
Defining $I_B = \left( \begin{array}{cccc} I & I & \ldots & I \end{array} \right)$ and
\begin{equation}
\resizemath{.5\hsize}{
\Gamma_l = \left( \begin{array}{ccc}
    \sqrt{\beta_0} G_l & 0 & \ldots \\
    0 & \sqrt{\beta_1} G_l  & \ldots \\
    \vdots & \vdots & \ddots
  \end{array} \right)
}
\end{equation}
allows further reduction to
$
\mu \norm{\sqrt{I_B (\Gamma_0 \mb{x})^2 + I_B (\Gamma_1 \mb{x})^2}}_1 \;.
$

Problem~\eq{convl1grdv} can be written in standard ADMM form as
\begin{align}
\hspace{-2mm}
\argmin_{\mb{x},\mb{y}_0,\mb{y}_1,\mb{y}_2} \; \frac{1}{2}
\normsz[\big]{D\mb{x} - \mb{s}}_2^2 &+ \lambda
\norm{\mbs{\alpha} \odot \mb{y}_2}_1 +
\mu \normsz[\Big]{\sqrt{I_B \mb{y}_0^2 + I_B \mb{y}_1^2}}_1 \nonumber \\
& \mkern-36mu  \text{ s.t. }
  \left( \begin{array}{c} \Gamma_0 \mb{x} \\ \Gamma_1 \mb{x} \\
     \mb{x} \end{array} \right)
  -
 \left( \begin{array}{c} \mb{y}_0 \\ \mb{y}_1 \\ \mb{y}_2 \end{array} \right)
 = 0
  \; .
\label{eq:cbpdnsplitmdv}
\end{align}
The resulting $\mb{x}$ subproblem has the same form as~\eq{cbpdngrdl1prob} and can be solved in the same way. The $\mb{y}_2$ subproblem is the same as~\eq{tvy2prob} and can be solved in the same way, while the $\mb{y}_0, \mb{y}_1$ subproblem, which only differs from~\eq{tvy01prob} in the first term, can be solved by
\vspace{-1.6mm}
\begin{align}
\mb{y}_l = \frac{\mb{z}_l}{\sqrt{I_B \mb{z}_0^2 + I_B \mb{z}_1^2}} \max\Big(0, \sqrt{I_B \mb{z}_0^2 + I_B \mb{z}_1^2} - \frac{\mu}{\rho}\Big)
\label{eq:l2shrnkvec}
\vspace{-2.8mm}
\end{align}
where $\mb{z}_l = \Gamma_l \mb{x} + \mb{u}_l$ for $l \in \{0,1\}$.

\subsection{Scalar TV in Image Domain}
\label{sec:rectv}

The use of TV regularization here is motivated as an exploration of additional or alternative forms of regularization to the standard $\ell_1$ regularization applied to the coefficient maps $\mb{x}$. An alternative way of introducing TV regularization, however, would be to consider it as a regularization on the components $D_m \mb{x}_m$ of the reconstructed image, which can be written as \vspace{-1mm}
\begin{multline}
\resizemath{.8\hsize}{
   \argmin_{\{\mb{x}_m\}}  \frac{1}{2} \normsz[\Big]{\sum_m \mb{d}_m \ast
     \mb{x}_m - \mb{s}}_2^2 + \lambda \sum_m \alpha_m \norm{\mb{x}_m}_1 +
  } \\[-1pt]
\resizemath{.8\hsize}{  \mu  \normsz[\bigg]{
 \sqrt{\vphantom{\sum} \smash[b]{\Big(\mb{g}_0 \ast \sum_m \beta_m \mb{d}_m \ast
   \mb{x}_m \Big)^2 + \Big(\mb{g}_1 \ast \sum_m  \beta_m \mb{d}_m \ast
   \mb{x}_m \Big)^2}}}_1
\; .
}
\label{eq:convl1grdidr}
\end{multline}

The final TV term can be expressed as
\vspace{-1mm}
\[
\resizemath{.8\hsize}{
\mu  \normsz[\bigg]{
 \sqrt{\vphantom{\sum} \smash[b]{\Big( \sum_m \beta_m (\mb{g}_0 \ast \mb{d}_m) \ast \mb{x}_m
   \Big)^2 \!\!+\!
  \Big( \sum_m \beta_m (\mb{g}_1 \ast \mb{d}_m) \ast
  \mb{x}_m\Big)^2}}}_1 \;.
}
\vspace{-1mm}
\]
Introducing linear operators $G_{l,m}$ defined such that $G_{l,m}
\mb{x} =  \beta_m (\mb{g}_l \ast \mb{d}_m) \ast \mb{x}$, this can be
written as
\vspace{-1mm}
\[
\resizemath{.65\hsize}{
 \mu  \normsz[\bigg]{ \sqrt{\vphantom{\sum} \smash[b]{\Big( \sum_m G_{0,m} \mb{x}_m \Big)^2 +
  \Big( \sum_m G_{1,m} \mb{x}_m \Big)^2 }}}_1 \;,
}
\vspace{-1mm}
\]
and defining
$\Gamma_l = \left( \begin{array}{ccc}
    G_{l,0} & G_{l,1}  & \ldots
  \end{array} \right)$
allows further reduction to
$
\mu \normsz[\big]{\sqrt{(\Gamma_0 \mb{x})^2 + (\Gamma_1 \mb{x})^2}}_1 \;.
$

\vspace{4pt}
Problem~\eq{convl1grdidr} can be written in standard ADMM form as
\vspace{-1mm}
\begin{align}
\hspace{-2mm}
\argmin_{\mb{x},\mb{y}_0,\mb{y}_1,\mb{y}_2} \frac{1}{2}
\normsz[\big]{D\mb{x} - \mb{s}}_2^2 &+ \lambda
\norm{\mbs{\alpha} \odot \mb{y}_2}_1 +
\mu \normsz[\Big]{\sqrt{\mb{y}_0^2 + \mb{y}_1^2}}_1 \nonumber \\[-2pt]
& \mkern-36mu  \text{ s.t. }
  \left( \begin{array}{c} \Gamma_0 \mb{x} \\ \Gamma_1 \mb{x} \\
     \mb{x} \end{array} \right)
  -
 \left( \begin{array}{c} \mb{y}_0 \\ \mb{y}_1 \\ \mb{y}_2 \end{array} \right)
 = 0
  \; .
\label{eq:cbpdnsplitmdid}
\end{align}
The resulting $\mb{x}$ subproblem corresponding to~\eq{bpdnxprob} has
the form \vspace{-1mm}
\begin{align}
   \argmin_{\mb{x}} & \frac{1}{2} \norm{D
     \mb{x} - \mb{s}}_2^2 +
   \frac{\rho}{2} \norm{\Gamma_0 \mb{x} - \mb{y}_0 + \mb{u}_0}_2^2  +
   \nonumber \\&
   \frac{\rho}{2} \norm{\Gamma_1 \mb{x} - \mb{y}_1 + \mb{u}_1}_2^2 +
   \frac{\rho}{2} \norm{\mb{x} - \mb{y}_2 + \mb{u}_2}_2^2 \; .
\label{eq:cbpdngrdl1probid}
 \vspace{-1mm}
\end{align}
and the solution of the equivalent DFT domain problem is given by
\vspace{-1mm}
\begin{align}
  (\hat{D}^H \hat{D} + &\rho I + \rho \hat{\Gamma}_0^H \hat{\Gamma}_0
  + \rho \hat{\Gamma}_1^H \hat{\Gamma}_1) \hat{\mb{x}} =
  \hat{D}^H \hat{\mb{s}} + \rho \left(\hat{\mb{y}}_2 - \hat{\mb{u}}_2
    + \vphantom{\hat{\Gamma}_0^H}  \right. \nonumber \\[-3pt]
 & \left.  \hat{\Gamma}_0^H (\hat{\mb{y}}_0 - \hat{\mb{u}}_0)   +
      \hat{\Gamma}_1^H (\hat{\mb{y}}_1 - \hat{\mb{u}}_1)
\right)   \; .
\label{eq:tvxprobid}
\vspace{-1mm}
\end{align}
Although the left hand side has the same algebraic form as that of~\eq{tvxprob}, here $\hat{\Gamma}_0^H \hat{\Gamma}_0$ and $\hat{\Gamma}_1^H \hat{\Gamma}_1$ are rank-one rather than diagonal, and can therefore not be grouped together with the $\rho I$ term as in the solution for~\eq{tvxprob}. In this case the left hand side is rank-three plus a diagonal: while it cannot be solved using the simple Sherman-Morrison approach, there is still an efficient solution via iterated application of the Sherman-Morrison formula, as used to solve the CBPDN problem for a multi-channel image and dictionary~\cite{wohlberg-2016-convolutional}. This involves a greater cost in terms of computation time, but there is a corresponding reduction in memory requirements because $\mb{y}_0$ and $\mb{y}_1$ are only of the size of the image rather than of the size of the set of coefficient maps.

The $\mb{y}$ subproblem for~\eq{cbpdnsplitmdid} has the same form as~\eq{tvy2prob} -- \eq{tvy01prob}, and can be solved in the same way.

\section{Results}
\label{sec:rslt}

\vspace{-2mm}

\begin{figure}[htbp]
  \centering \small
  \begin{tabular}{ccc}
    \subfigure[\label{fig:tstimg0}\protect\rule{0pt}{1.5em}
    Image 1]
               {\includegraphics[width=2.4cm]{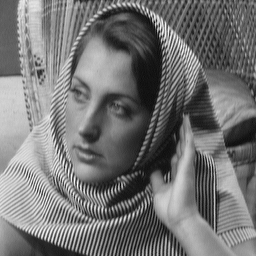}} &
    \hspace{-2mm}
    \subfigure[\label{fig:tstimg1}\protect\rule{0pt}{1.5em}
    Image 2]
               {\includegraphics[width=2.4cm]{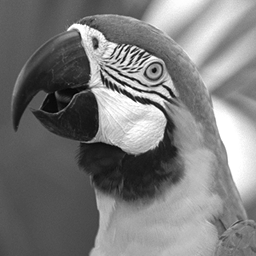}}
&
    \hspace{-2mm}
    \subfigure[\label{fig:tstimg2}\protect\rule{0pt}{1.5em}
    Image 3]
               {\includegraphics[width=2.4cm]{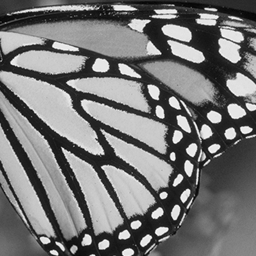}}
\\
    \hspace{-2mm}
    \subfigure[\label{fig:tstimg3}\protect\rule{0pt}{1.5em}
    Image 4]
               {\includegraphics[width=2.4cm]{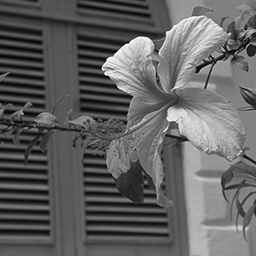}}
&
    \hspace{-2mm}
    \subfigure[\label{fig:tstimg4}\protect\rule{0pt}{1.5em}
    Image 5]
               {\includegraphics[width=2.4cm]{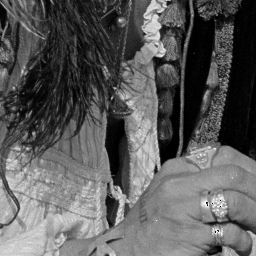}}
  \end{tabular}
  \vspace{-3mm}
  \caption{Set of $256 \times 256$ pixel noise-free test images.}
  \label{fig:tstimg}
\end{figure}

The performance of standard block-based sparse coding and the different convolutional sparse coding methods described in Sections~\ref{sec:cbpdn} and~\ref{sec:tvreg} was compared on a Gaussian white noise restoration problem. The standard sparse coding was computed via the Basis Pursuit DeNoising (BPDN) problem (\ie problem~\eq{convbpdnblk} where $D$ is a standard dictionary matrix) and the resulting denoised blocks were aggregated via averaging (weighted by the number of blocks covering each pixel) to obtain a denoised image.

Two different dictionaries, one standard and one convolutional, were learned from the same set of ten training images (selected from images on Flickr with a Creative Commons license) of $1024 \times 1024$ pixels each. The convolutional dictionary consisted of 128 filters of size $8 \times 8$, and was learned via the convolutional dictionary learning algorithm described in~\cite{wohlberg-2016-efficient}, while the standard dictionary consisted of 128 vectors of 64 coefficients each (\ie a vectorised $8 \times 8$ image block), and was learned via a non-convolutional variant of the algorithm used for learning the convolutional dictionary, applied to all $8 \times 8$ image blocks in the training images. The standard dictionary was used for the BPDN experiments and the convolutional dictionary was used for all CBPDN experiments.

A set of five greyscale reference images, depicted in~\fig{tstimg}, was constructed by cropping regions of $256 \times 256$ pixels from well-known standard test images. The regions were chosen to contain diversity of content while avoiding large smooth areas, and the size was chosen to be relatively small so that it would be computationally feasible to optimise method parameters via a grid search. The reference images were scaled so that pixel values were in the interval $[0,1]$, and corresponding test images were constructed by adding Gaussian white noise with a standard deviation of 0.05. Following standard practice~\cite{zeiler-2011-adaptive}\cite[Sec. 3]{wohlberg-2016-convolutional2}, the CBPDN decomposition was applied to highpass filtered images, obtained by subtracting a lowpass component computed by Tikhonov regularization~\cite[pg. 3]{wohlberg-2017-sporco} with regularization parameter $\lambda_{L} = 2.0$.

{\small
\begin{table}[htbp]
  \centering
  \begin{tabular}{|l|r|r|r|r|r|} \hline
   & \multicolumn{5}{|c|}{Test Image} \\ \hline
Method       &  1     & 2     & 3     & 4     & 5   \\ \hline\hline
BPDN         &  29.47 & 32.91 & \textit{30.08} & 31.73 & 30.19 \\ \hline\hline
CBPDN        &  29.31 & 32.70 & 29.76 & 31.27 & 30.09 \\ \hline
CBPDN + Grd  &  29.28 & 32.76 & 30.02 & 31.22 & 30.12 \\ \hline
CBPDN + STV  &  \textbf{30.17} & 33.01 & 29.90 & \textbf{32.09} &
\textbf{30.34} \\ \hline
CBPDN + VTV  &  29.60 & \textbf{33.04} & \textbf{29.96} & 31.63 & 30.31 \\
\hline
CBPDN + RTV  &  29.28 & 32.84 & 29.76 & 31.29 & 30.19 \\ \hline
 \end{tabular} \vspace{1mm}
 \caption{Comparison of denoising performance (PSNR in dB) of the different denoising methods for each of the five test images, with parameters individually optimised for each image. Bold values indicate the best performing CBPDN method. An italic value in the BPDN row indicates that BPDN gave the best overall performance for that image.}
  \label{tbl:cmpmain}
\vspace{-3mm}
\end{table}
}

For the first set of experiments, the results of which are displayed in~\tbl{cmpmain}, the denoising performance of the different methods was individually optimised for each image via a search over a logarithmically spaced grid on the $\lambda$ and $\mu$ parameters. The main points worth noting are:
 \begin{itemize}
 \setlength{\itemsep}{0.6pt}
 \item BPDN is consistently better than CBPDN by a small margin.
 \item CBPDN + Grd ($\ell_2$ of gradient regularization, as in~\cite[Sec. 4]{wohlberg-2016-convolutional2}) gives very similar performance to CBPDN, being slightly better on some test images and slightly worse on others.
 \item CBPDN + STV (see~\sctn{sclrtv}) gives the best overall performance on three of the five test images, with performance within a few tenths of a dB of the best in the other cases. It is consistently better than CBPDN, and better than BPDN in all but one of the test cases.
 \item In a comparison between CBPDN + STV and CBPDN + VTV (see~\sctn{vectv}), the former is sometimes better by a moderate margin, but when it is worse this is by a very small amount.
 \item CBPDN + RTV (see~\sctn{rectv}) is always worse than the other two TV-augmented CBPDN methods, and is sometimes no better than CBPDN.
 \end{itemize}
 The computation times per iteration for the different methods were approximately 0.5 s for BPDN and CBPDN, 0.6 s for CBPDN + Grd, 2.2 s for CBPDN + STV and CBPDN + VTV, and 2.4 s for CBPDN + RTV, \ie the improved performance of the TV methods is obtained at a significant computational cost.

{\small
\begin{table}[htbp]
  \centering
  \begin{tabular}{|l|r|r|r|r|r|} \hline
   & \multicolumn{5}{|c|}{Test Image} \\ \hline
Method       &  1     & 2     & 3     & 4     & 5   \\ \hline \hline
CBPDN + Grd  &  -2.31 & -3.16  &  -2.51 & -1.39 & -0.94 \\ \hline
CBPDN + STV  &  +0.04 &  -0.22 &  -0.04 & -0.03 & +0.03  \\ \hline
CBPDN + VTV  &  -0.64 &  -0.77 &  -0.89 & -0.29 & -0.34 \\ \hline
CBPDN + RTV  &  -1.28 &  -0.66 &  -0.73 & -0.47 & -0.33 \\ \hline
 \end{tabular} \vspace{1mm}
 \caption{PSNR difference in dB between results for optimisation over
   both $\lambda$ and $\mu$ (\tbl{cmpmain}) and for optimisation over
   $\mu$ only, with $\lambda = 0$.}
  \label{tbl:cmplr0dif}
\end{table}
}

The second set of experiments evaluated the efficacy of the terms augmenting plain CBPDN by comparing the denoising performance at the best choices of both $\lambda$ and $\mu$ (as in~\tbl{cmpmain}) with the same method with $\lambda$ fixed to zero and optimisation only over $\mu$. (There is no need to perform a corresponding comparison with $\mu$ fixed to zero since this corresponds to the baseline CBPDN method.) The differences between the PSNR values of the methods optimised over both parameters and only optimised over $\mu$ are displayed in~\tbl{cmplr0dif}. Note that, for CBPDN + STV, there is a positive difference in two cases and a very small negative difference in two other cases, \ie for most of the test images, the convolutional representation with only a TV regularization term is competitive with the baseline CBPDN. For all of the other methods the performance is substantially degraded without the $\ell_1$ term.

{\small
\begin{table}[htbp]
  \centering
  \begin{tabular}{|l|r|r|r|r|r|} \hline
   & \multicolumn{5}{|c|}{Test Image} \\ \hline
Method       &  1     & 2     & 3     & 4     & 5   \\ \hline\hline
BPDN         &  29.47 & 32.03 & \textit{29.92} & 31.38 & 30.19 \\ \hline\hline
CBPDN        &  29.24 & 31.73 & 29.54 & 30.89 & 30.00 \\ \hline
CBPDN + STV  &  \textbf{29.90} & \textbf{32.36} & \textbf{29.86} &
 \textbf{31.68} & \textbf{30.29} \\ \hline
CBPDN + VTV  &  29.54 & 32.35 & \textbf{29.86} & 31.34 & 30.25 \\ \hline
CBPDN + RTV  &  29.16 & 32.49 & 29.76 & 31.25 & 30.19 \\ \hline
 \end{tabular} \vspace{1mm}
 \caption{Comparison of denoising performance (PSNR in dB) of the different denoising methods for each of the five test images, all with the same parameters obtained by optimising over a separate image set. Bold values indicate the best performing CBPDN method. An italic value in the BPDN row indicates that BPDN gave the best overall performance for that image.}
  \label{tbl:cmpfxpr}
\end{table}
}

The final set of experiments considers a more realistic scenario in which ground truth is not available for parameter selection for the test images, making it necessary to choose the $\lambda$ and $\mu$ parameters by optimising over a distinct parameter selection image set. The same $\lambda$ and $\mu$ parameters were selected for all test images by finding the values giving the best average performance for a separate image set, again via a search on a logarithmically spaced grid. The results for this experiment are presented in~\tbl{cmpfxpr}. Overall, the relative performances of the different methods do not differ qualitatively from those of the experiments reported in~\tbl{cmpmain}. (CBPDN + Grd is excluded from this set of experiments since it is clear from the first two sets of experiments that it is not competitive.)

\section{Conclusions}
\label{sec:cnclsn}

While a strictly apples-to-apples comparison between BPDN and CBPDN denoising methods is difficult to construct, the careful attempt reported here indicates that BPDN is slightly superior to baseline CBPDN, but that augmentation of the baseline CBPDN functional with the appropriate TV term substantially boosts performance, surpassing that of BPDN in all but one of the five test cases considered here. With respect to the specific form of additional TV term, scalar TV applied independently to each coefficient map is somewhat superior to a joint vector TV term over all of the coefficient maps, and both of these methods are substantially superior to TV applied in the reconstruction domain rather than to the coefficient maps, indicating that the gain from a TV term on the coefficient maps should not be viewed simply as resulting from denoising via a synthesis of sparse representation and TV image models. It is particularly interesting that the convolutional sparse coding problem with only an STV penalty is competitive in performance with the usual CBPDN form with only an $\ell_1$ penalty. At a more abstract level, these results suggest that penalties that exploit the spatial structure of the coefficient maps are necessary to achieve the true potential of the convolutional model.

Implementations of the algorithms proposed here are included in the Python version of the SPORCO library~\cite{wohlberg-2016-sporco, wohlberg-2017-sporco}.

\bibliographystyle{IEEEtranD}
\bibliography{cbpdntv}

\end{document}